\documentclass[journal]{IEEEtran}
\ifCLASSINFOpdf
\else
\fi
\usepackage{cite}
\usepackage{amsmath,amssymb,amsfonts}
\usepackage{algorithmic}
\usepackage{algorithm}
\usepackage{epsfig}
\usepackage{epstopdf}
\usepackage{graphicx}
\usepackage{textcomp}
\usepackage{xcolor}

\begin{document}
%
\title{A Complete Discriminative Tensor Representation Learning for Two-Dimensional Correlation Analysis}
%
%
%

\author{Lei Gao,~\IEEEmembership{Member,~IEEE,}
        Ling Guan,~\IEEEmembership{Fellow,~IEEE}
\thanks{L. Gao and L. Guan are with the Department of Electrical, Computer and Biomedical Engineering, Ryerson University, Toronto, ON M5B 2K3, Canada (email:iegaolei@gmail.com; lguan@ee.ryerson.ca).}}
\maketitle

\begin{abstract}
As an effective tool for two-dimensional data analysis, two-dimensional canonical correlation analysis (2DCCA) is not only capable of preserving the intrinsic structural information of original two-dimensional (2D) data, but also reduces the computational complexity effectively. However, due to the unsupervised nature, 2DCCA is incapable of extracting sufficient discriminatory representations, resulting in an unsatisfying performance. In this letter, we propose a complete discriminative tensor representation learning (CDTRL) method based on linear correlation analysis for analyzing 2D signals (e.g. images). This letter shows that the introduction of the complete discriminatory tensor representation strategy provides an effective vehicle for revealing and extracting the discriminant representations across the 2D data sets, leading to improved results. Experimental results show that the proposed CDTRL outperforms state-of-the-art methods on the evaluated data sets.
\end{abstract}


%
\IEEEpeerreviewmaketitle

\section{Introduction}
%
%
%

\IEEEPARstart{R}{apid} developments in sensory and computing technology have enabled the accessibility of multiple data/information sources representing the same phenomenon from a variety of acquisition techniques and devices. Specifically, multiple data/information sources in images have been playing a vital and central role in two and multidimensional signal processing. Therefore, multiple data/information representation learning is becoming a challenging but increasingly significant research topic in the signal processing and statistics communities [1-2, 21].\\\indent Recently, correlation analysis has drawn more attention in academic and industrial sectors for multiple data/information representation learning [3, 26]. The aim of correlation analysis is to measure and evaluate intrinsic correlation across different data sets. As a typical representation for linear correlation analysis, canonical correlation analysis (CCA) plays important roles and has been applied to signal analysis, visual representation and other tasks [4-5]. In CCA, not only is the correlation taken into consideration, but the canonical characteristic is employed to gain maximal correlation. Consequentially, CCA is widely utilized for cross-modal correlation analysis, such as audiovisual-based emotion recognition [6] and medical imaging analysis [7], etc. Nevertheless, in the field of image and visual computing, 2D data samples are reshaped into one-dimensional vectors before CCA is performed. It is known that the reshaping operation breaks the spatial structural information of 2D data sets and introduces a higher computing complexity [8].\\\indent Therefore, two-dimensional CCA (2DCCA) is presented to address this problem [9]. 2DCCA aims to identify the maximal linear correlation among 2D data sets directly, without going through the reshaping operation. In addition, since the size of covariance matrices for 2D data is smaller than the reshaped vector-based covariance matrices, 2DCCA leads to a lower computational complexity than CCA [9,18]. Assume we have $N$ two dimensional samples with a size of $m \times n$. Then the computational complexity of the traditional CCA is on the order \emph{O$((mn)^{3})$} while 2DCCA only requires a computational complexity of \emph{O$((m)^{3})$} or \emph{O$((n)^{3})$}. Moreover, local two-dimensional canonical correlation analysis (L2DCCA) [17] and two-directional two-dimensional kernel canonical correlation analysis ($(2D)^2KCCA$) [10] are proposed as an extension of 2DCCA. In L2DCCA, the local structural information is introduced to the 2DCCA space, revealing more useful representations between 2D data sets with the computational complexity on the order of \emph{$O((m)^{3}+(N)^{2}m)$} or \emph{$O((n)^{3}+(N)^{2}n)$}. The main purpose of $(2D)^2KCCA$ is to explore the nonlinear correlation between different 2D data sets, in order to achieve better classification performance. However, as far as we know, since most of the existing 2DCCA and related algorithms belong to the unsupervised learning category, they are not able to measure and extract sufficient discriminatory representations across 2D data sets effectively, resulting in an unsatisfying recognition performance. To address the aforementioned issues, a complete discriminant tensor representation learning (CDTRL) is proposed.\\\indent The contributions of this letter are summarized as follows: 1) A discriminant tensor representation learning solution is proposed to explore discriminative representations from 2D data sets. 2) The discriminative representations derived from the range space and the null space of the within-class matrix are utilized jointly to construct a complete discriminant descriptor for 2D correlation analysis. 3) The generality of the proposed CDTRL is validated by two examples. This generic nature guarantees that CDTRL can be used in a broad range of applications.\\\indent The remainder of this letter is organized as follows: A review of related work is presented in Section II. The proposed CDTRL is formulated in Section III. Experimental results and analysis are shown in Section IV. Conclusions are given in Section V.
\section{Related Work}
In this section, we briefly introduce the fundamentals of the 2DCCA and L2DCCA methods, respectively.
\subsection{2DCCA}
Suppose we have two 2D data sets $X$ and $Y$. The samples from $X$ and $Y$ are defined as $ {X_i} \in {R^{m \times n}},{Y_i} \in {R^{p \times q}} (i=1,2,...N)$, where $N$ is the number of samples. The mean matrices of $X$ and $Y$ are calculated as below
\begin{small}
\begin{equation}
{M_X} = 1/N\sum\limits_{i = 1}^N {{X_i}},
{M_Y} = 1/N\sum\limits_{i = 1}^N {{Y_i}},
\end{equation}
\end{small}
and denoted as $\mathop X^{\sim} = X - {M_X} $ and $\mathop Y^{\sim} = Y - {M_Y} $. Then, the purpose of 2DCCA is to find left projected matrices ${L_X}$ \& ${L_Y}$, and right projected matrices ${R_X}$ \& ${R_Y}$, which maximize the correlation between ${L_X}'X^{\sim}R_X$ and ${L_Y}'Y^{\sim}R_Y$ jointly. The optimization of 2DCCA is given in equation (2) [9]
\begin{equation}
\begin{array}{l}
 \arg \max {\mathop{\rm}} ({L_X}^\prime {X^{\sim}}{R_X} \cdot ({L_Y}^\prime {Y^{\sim}}{R_Y})^\prime), \\
 \\
 s.t.{\mathop{\rm var}} ({L_X}^\prime {X^{\sim}}{R_X}) = {\mathop{\rm var}} ({L_Y}^\prime {Y^{\sim}}{R_Y}) = 1, \\
 \end{array}
\end{equation}
where `var' denotes the variance of a given variable. In equation (2), we are capable of finding the solutions to ${L_X}$, ${L_Y}$, ${R_X}$ and ${R_Y}$ by a generalized eigen-value (GEV) algorithm.
\subsection{L2DCCA}
For the L2DCCA method, a manifold algorithm is introduced to 2DCCA to explore the local structural information between the 2D data sets. The weight ${A^X}_{ij}$ between two samples $X_i$ and $X_j$ in $X$ is formulated as
\begin{small}
\begin{equation}
{A^X}_{ij} = \exp (-\frac{{{{\left\| {{X_i} - {X_j}} \right\|}^2_{F}}}}{{{\sigma ^2}}}),
\end{equation}
\end{small}
where ${{{{\left\| {.} \right\|}_{F}}}}$ is the Frobenius norm. The aim of L2DCCA is to find two pairs of projected matrices ${L_{K_X}}$, ${L_{K_Y}}$, ${R_{K_X}}$ and ${R_{K_Y}}$ according to the relation in (4)
\begin{equation}
\begin{array}{l}
 \arg \max {\mathop{\rm}} ({A^X}_{ij}{L_{K_X}}^\prime X^{\sim} {R_{K_X}} \cdot ({A^Y}_{ij}{L_{K_Y}}^\prime Y^{\sim}{R_{K_Y}})^\prime), \\
 \\
 s.t.{\mathop{\rm var}} ({A^X}_{ij}{L_{K_X}}^\prime X^{\sim} {R_{K_X}}) = {\mathop{\rm var}} ({A^Y}_{ij}{L_{K_X}}^\prime {Y^{\sim}} {R_{K_Y}}) = 1. \\
 \end{array}
\end{equation}
Then, solutions to equation (4) are obtained by using the GEV algorithm iteratively.
\section{The Proposed CDTRL Method}
Given two sets of 2D data $X = [{X_1},{X_2},...{X_N}]$ and $Y = [{Y_1},{Y_2},...{Y_N}]$, where $ {X_i} \in {R^{m \times n}},{Y_i} \in {R^{p \times q}} (i=1,2,...N)$. Then, the zero-mean data sets are denoted as $X^{\sim}$ and $Y^{\sim}$, respectively. Assume the size of the two paired project matrices in 2DCCA satisfies the relation: $ L_X \in {R^{m \times {d_1}}}, L_Y \in {R^{p \times {d_1}}}$, $ R_X \in {R^{n \times {d_2} }}$ and $R_Y \in {R^{q \times {d_2} }}$. Then the projections of 2DCCA on the two 2D data sets are expressed as follows
\begin{equation}
 {X_{P}} = {L_X}^\prime {X^{\sim}}{R_X}, {Y_{P}} = {L_Y}^\prime {Y^{\sim}}{R_Y},
\end{equation}
where ${X_{P}} \in {R^{d_1 \times d_2 \times N}}$ and ${Y_{P}} \in {R^{d_1 \times d_2 \times N}}$. Let $n_k$ be the number of samples in the \emph{k}th class and satisfy the following relation
\begin{small}
\begin{equation}
\sum\limits_{k = 1}^c {{n_k} = N},
\end{equation}
\end{small}
where $c$ is the number of classes in $X$ and $Y$.\\\ From equation (5), it is known that there are $N$ paired maps in the ${d_1} \times {d_2}$ plane and their correlation achieves maximum on this plane, which is depicted in Figure 1.\\
\centerline{\includegraphics[height=0.6in,width=1.5in]{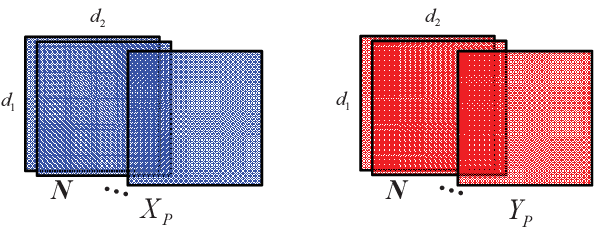}}\\ {Figure. 1 The representtaion of the $N$ paired maps ($X_P$, $Y_P$). }\\

According to Figure. 1, it is observed that $X_P$ and $Y_P$ can be represented by two three-dimensional tensors. Our aim is to find the discriminative representations between the two three-dimensional tensors. However, it is an interesting but challenging research topic to extract the discriminative representations between high dimensional tensors [22-23]. To address this issue, a discriminant tensor strategy is proposed according to the canonical property. According to the canonical property, the correlation satisfies the relation in equation (7)
\begin{equation}
\left\{ \begin{array}{l}
 {\rm{}}({L_{{X}}}^\prime {X^\sim_u}{R_{{X}}} \cdot ({L_{{Y}}}^\prime {Y^\sim_w}{R_{{Y}}})^\prime) = 0, \\
 {\rm{}}({L_{{X}}}^\prime {X^\sim_w}{R_{{X}}} \cdot ({L_{{Y}}}^\prime {Y^\sim_u}{R_{{Y}}})^\prime) = 0, \\
 \end{array} \right.
\end{equation}
where $u,w \in [1,2,...N]$. Therefore, the correlation between different paired maps is 0. Then, the $N$ paired maps ($X_P$, $Y_P$) can be connected as shown in Figure. 2 with the maximum correlation.\\
\centerline{\includegraphics[height=0.8in,width=0.8in]{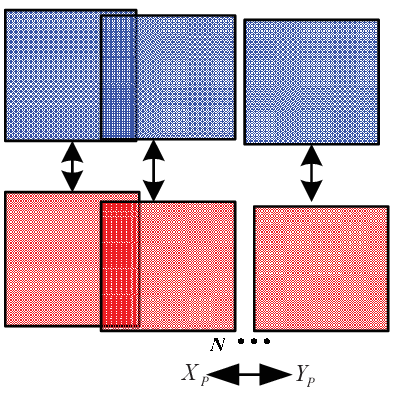}}\\ {Figure. 2 The connection of the $N$ paired maps ($X_P$, $Y_P$).}\\\indent

The connected $N$ paired maps are expressed mathematically in equation (8)
\begin{small}
\begin{equation}
F = \left[ \begin{array}{l}
 {X_{P}} \\
 {Y_{P}} \\
 \end{array} \right] = \left[ \begin{array}{l}
 {L_X}^\prime {X^{\sim}}{R_X} \\
 {L_Y}^\prime {Y^{\sim}}{R_Y} \\
 \end{array} \right],
\end{equation}
\end{small}
or in matrix-vector form
\begin{small}
\begin{equation}
\begin{array}{l}
 F = {\left[ {\left( {\begin{array}{*{20}{c}}
   {{L_X}} & {\bf{0}}  \\
   {\bf{0}} & {{L_Y}}  \\
\end{array}} \right)} \right]^{'}}\left[ {\left( {\begin{array}{*{20}{c}}
   {{X^{\sim}}} & {\bf{0}}  \\
   {\bf{0}} & {{Y^{\sim}}}  \\
\end{array}} \right)} \right]\left[ \begin{array}{l}
 {R_X} \\
 {R_Y} \\
 \end{array} \right]. \\
 \end{array}
\end{equation}
\end{small}
Thus, discriminative representation extraction from tensors is accomplished by maximizing the between-class matrix and minimizing the within-class matrix from $N$ connected maps jointly.\\\indent The total mean matrix of $F$ is calculated in equation (10)
\begin{small}
\begin{equation}
{M_F} = \frac{1}{N}\sum\limits_{i = 1}^N {{F_i}},
\end{equation}
\end{small}
and the mean matrix of the $j$th class in $F$ is calculated in (11)
\begin{small}
\begin{equation}
{M_{{F_j}}} = \frac{1}{{{n_j}}}\sum\limits_{s = 1}^{{n_j}} {{F_{js}}},
\end{equation}
\end{small}
where $F_{js}$ represents the \emph{s}th sample in class $j$.\\\indent Afterwards, the within-class and between-class matrices of $N$ samples in ${d_1} \times {d_2}$ space are computed using the following formulas
\begin{small}
\begin{equation}
\begin{array}{l}
 {S_w} = \sum\limits_{j = 1}^c {\sum\limits_{s \in j} {({F_{js}} - {M_F}){{({F_{js}} - {M_F})}^{'}}} },  \\
 {S_b} = \sum\limits_{j = 1}^c {{n_j}({F_j} - {M_{{F_j}}}){{({F_j} - {M_{{F_j}}})}^{'}}}.  \\
 \end{array}
\end{equation}
\end{small}
Then, CDTRL aims to find the optimal left projected matrix $l$ and right projected matrix $r$ as formulated in equation (13)
\begin{small}
\begin{equation}
\begin{array}{l}
 {S_{b,lr}} = \sum\limits_{j = 1}^c {{n_j}{l^{'}}({F_j} - {M_{{F_j}}})r{r^{'}}{{({F_j} - {M_{{F_j}}})}^{'}}l},  \\
 {S_{w,lr}} = \sum\limits_{j = 1}^c {\sum\limits_{s \in j} {{l^{'}}({F_{js}} - {M_F})r{r^{'}}{{({F_{js}} - {M_F})}^{'}}l} }.  \\
 \end{array}
\end{equation}
\end{small}
The proposed CDTRL method is formulated as the following optimization problems to extract the complete discriminative representations corresponding to joint utilization of the range space (14) and the null space (15) of the within-class matrix [19]
\begin{equation}
\begin{array}{l}
 \arg \mathop {\max }\limits_{l,r} \frac{{tr({S_{b,lr}})}}{{tr({S_{w,lr}})}} \\
  = \arg \mathop {\max }\limits_{l,r} \frac{{tr(\sum\limits_{j = 1}^c {{n_j}{l^{'}}({F_j} - {M_{{F_j}}})r{r^{'}}{{({F_j} - {M_{{F_j}}})}^{'}}l} )}}{{tr(\sum\limits_{j = 1}^c {\sum\limits_{s \in j} {{l^{'}}({F_{js}} - {M_F})r{r^{'}}{{({F_{js}} - {M_F})}^{'}}l} } )}} \\
 s.t.\quad{\rm{ }}{S_{w,lr}} \ne {\bf{0}}, \\
 \end{array}
\end{equation}
or
\begin{small}
\begin{equation}
\begin{array}{l}
 \arg \mathop {\max }\limits_{l,r} tr({S_{b,lr}}) \\
  = \arg \mathop {\max }\limits_{l,r} tr(\sum\limits_{j = 1}^c {{n_j}{l^{'}}({F_j} - {M_{{F_j}}})r{r^{'}}{{({F_j} - {M_{{F_j}}})}^{'}}l} )\\
 s.t.\quad{\rm{ }}{S_{w,lr}} = {\bf{0}}, \\
 \end{array}
\end{equation}
\end{small}
where $tr$ denotes the trace operation of a matrix. Since $l$ \& $r$ are two independent matrices and there is no intrinsic relation between them, it is difficult to compute the optimal $l$ \& $r$ simultaneously [20]. To obtain the optimal $l$ and $r$, an iterative strategy is proposed in this letter, which is described as follows. Given an initial value of $r$ (or $l$) and the projected matrix of $l$ (or $r$) is obtained by solving the optimization problem in equation (14) or (15). Afterwards, the projected matrix $r$ (or $l$) is updated with the previous $l$ (or $r$). Thus, $l$ and $r$ are determined by iteratively calculating the ratio between $tr({S_{b,lr}})$ and $tr({S_{w,lr}})$ or $tr({S_{b,lr}})$ until convergence. Then, the computational complexity of CDTRL is on the order of \emph{O$((m)^{3}+(N)^{3})$} or \emph{O$((n)^{3}+(N)^{3})$}. A more detailed description is summarized in the following subsections.
\begin{figure*}[t]
\centering
\includegraphics[height=0.7in,width=5.5in]{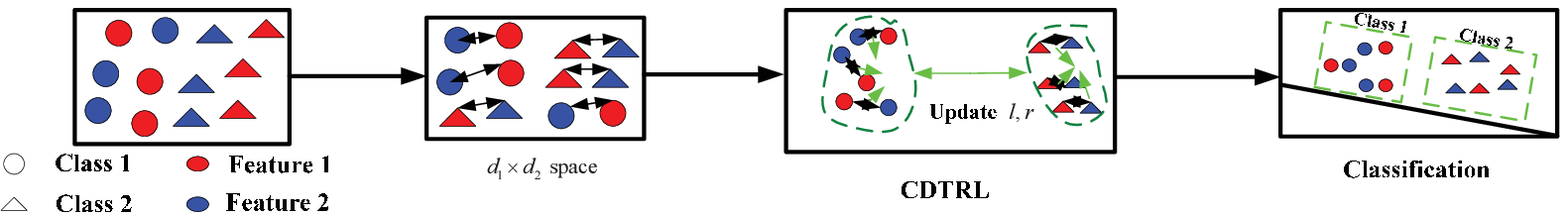}\\ Figure. 3 The diagram of the proposed CDTRL\\
\end{figure*}
\subsection{Update the optimal matrix $l$}
For a given $r$, $S_{w, lr}$ and $S_{b, lr}$ are formulated in (16)
\begin{small}
\begin{equation}
\begin{array}{l}
 {S_{w, lr}} = {l^{'}}{S_{w, lr}}^rl, \\
 {S_{b, lr}} = {l^{'}}{S_{b, lr}}^rl, \\
 \end{array}
\end{equation}
\end{small}
where
\begin{small}
\begin{equation}
\begin{array}{l}
 {S_{w, lr}}^r = \sum\limits_{j = 1}^c {\sum\limits_{s \in j} {({F_{js}} - {M_F})r{r^{'}}{{({F_{js}} - {M_F})}^{'}}} },  \\
 {S_{b, lr}}^r = \sum\limits_{j = 1}^c {{n_j}({F_j} - {M_{{F_j}}})r{r^{'}}{{({F_j} - {M_{{F_j}}})}^{'}}}.  \\
 \end{array}
\end{equation}
\end{small}
Then, using the Lagrange multiplier method, the projected matrix $l$ is obtained by solving the following optimization functions
\begin{small}
\begin{equation}
\begin{array}{l}
{S_{w,lr}}^rl = \lambda {S_{b,lr}}^rl\\
s.t.\quad{\rm{ }}{S_{w,lr}} \ne {\bf{0}}, \\
\end{array}
\end{equation}
\end{small}
or
\begin{small}
\begin{equation}
\begin{array}{l}
\arg \mathop {\max }\limits_l tr({l^{'}}{S_{b,lr}}^rl)\\
s.t.\quad{\rm{ }}{S_{w,lr}} = {\bf{0}}. \\
\end{array}
\end{equation}
\end{small}
\subsection{Update the optimal matrix $r$}
Similarly, with a given $r$, $S_{w, lr}$ and $S_{b, lr}$ are formulated in (20)
\begin{small}
\begin{equation}
\begin{array}{l}
 {S_{w, lr}} = {r^{'}}{S_{w, lr}}^lr, \\
 {S_{b, lr}} = {r^{'}}{S_{b, lr}}^lr, \\
 \end{array}
\end{equation}
\end{small}
where
\begin{small}
\begin{equation}
\begin{array}{l}
 {S_{w, lr}}^l = \sum\limits_{j = 1}^c {\sum\limits_{s \in j}{({F_{js}} - {M_F})l{l^{'}}{{({F_{js}} - {M_F})}^{'}}} },  \\
 {S_{b, lr}}^l = \sum\limits_{j = 1}^c {{n_j}({F_j} - {M_{{F_j}}})l{l^{'}}{{({F_j} - {M_{{F_j}}})}^{'}}}.  \\
 \end{array}
\end{equation}
\end{small}
Again, the projected matrix $r$ is obtained by solving the optimization functions in (22) or (23)
\begin{small}
\begin{equation}
\begin{array}{l}
{S_{w,lr}}^lr = \eta {S_{b,lr}}^lr\\
s.t.\quad{\rm{ }}{S_{w,lr}} \ne {\bf{0}}, \\
\end{array}
\end{equation}
\end{small}
or
\begin{small}
\begin{equation}
\begin{array}{l}
\arg \mathop {\max }\limits_l tr({r^{'}}{S_{b,lr}}^lr)\\
s.t.\quad{\rm{ }}{S_{w,lr}} = {\bf{0}}. \\
\end{array}
\end{equation}
\end{small}
The CDTRL algorithm is summarized in \textbf{Algorithm 1} and the block diagram is depicted in Figure. 3.
\begin{tiny}
\begin{algorithm}[]
\caption{The proposed CDTRL algorithm}
\label{alg:Framwork}
\begin{algorithmic}
\REQUIRE ~~\\
\textbf{*} Given two 2D data sets X and Y with $N$ samples.\\
\ENSURE ~~\\
\STATE \textbf{*} Compute the zero-mean data sets $\mathop X^{\sim}$ and $\mathop Y^{\sim}$.
\STATE \textbf{*} Find projected matrices ${L_X}$, ${L_Y}$, ${R_X}$ and ${R_Y}$ according to equation (2).\\
\STATE \textbf{*} Construct the matrix $F$ based on equation (9).\\
\STATE \textbf{*} Calculate the matrices ${S_{w, lr}}^r$, ${S_{b, lr}}^r$, ${S_{w, lr}}^l$ and ${S_{b, lr}}^l$, respectively.
\STATE \textbf{*} Calculate the $r$ and $l$ according to (18)(22) or (19)(23) until convergence iteratively.
\RETURN the optimal matrices $l$ and $r$.
\end{algorithmic}
\end{algorithm}
\end{tiny}
\section{Experimental Results and Analysis}
To examine the performance of CDTRL, we conduct experiments on AR [11] and FERET [12] face data sets, respectively. The 2D samples in AR database are collected under various conditions, such as different facial emotions, lighting. etc. To verify the generality of the proposed method, 480 samples of 120 subjects are chosen randomly from the AR face database and each of them was normalized to a size of 50 $\times$ 50 pixels. In addition, for each subject, we select four different face images, including one reference sample and the other three samples under different expressions and illumination conditions. 
In the FERET database, 600 samples of 200 subjects with a size of 80 $\times$ 80 pixels are selected. For each person, it contains three samples with different poses (e.g. front, left and right). 
In what follows, we will test the performance of CDTRL under different conditions (e.g. facial expressions and illumination) on the AR database and various facial pose images on the FERET database.

\subsection{Experiments under Different Expressions and Illumination on the AR Database}
In this subsection, we will conduct experiments with the proposed CDTRL and then compare the performances of CCA [5], 2DCCA [9], principal component analysis (PCA) [13], two-dimensional PCA (2DPCA) [14], linear discriminant analysis (LDA) [15], two-dimensional LDA (2DLDA) [16], Local 2DCCA (L2DCCA) [17], discriminative CCA (DCCA) [24] and labeled CCA (LCCA) [25]. Note, for CCA and related algorithms (such as 2DCCA, L2DCCA, DCCA, LCCA and CDTRL), we divide the chosen images into two groups. The first group contains only the reference samples while the remaining images are in the second group. As a result, 360 reference samples (three copies of 120 reference samples to match the samples in the second group) are in the first group $X$ and 360 samples with various conditions are stored in the second group $Y$.\\\indent Since the operations of PCA, LDA, 2DPCA and 2DLDA algorithms do not involve correlation analysis, these methods are either applied directly to the 480 chosen samples (2DPCA and 2DLDA), or to the samples reshaped into one-dimensional vectors (PCA and LDA). For the correlation based methods, 2DCCA, L2DCCA and CDTRL are performed on the 2D data sets X and Y directly while CCA, DCCA and LCCA work on samples reshaped into one-dimensional vectors. To further validate the effectiveness of the proposed method, the leave-one-out cross-validation strategy is utilized and recognition accuracies are tabulated in TABLE I. Viewing the table, it is evident that CDTRL yields performance superior to the others.
\vspace*{-10pt}
\begin{table}[h]
\normalsize
\renewcommand{\arraystretch}{0.75}
\caption{\normalsize{The recognition accuracy with different methods on the AR dataset}}
\setlength{\abovecaptionskip}{0pt}
\setlength{\belowcaptionskip}{10pt}
\centering
\tabcolsep 0.1in
\begin{tabular}{cc}
\hline
\hline
Method & Recognition Accuracy\\
\hline
PCA [13]  &93.33\%\\
2DPCA [14]  &94.17\%\\
LDA [15]  &95.00\%\\
2DLDA [16]  &95.83\%\\
CCA [5]  &95.83\%\\
DCCA [24]  &98.33\%\\
LCCA [25]  &96.67\%\\
2DCCA [9]  &97.50\%\\
L2DCCA [17]  &98.01\%\\
\textbf{The proposed CDTRL}  &\textbf{100.00\%}\\
\hline
\hline
\end{tabular}
\end{table}
\subsection{Experiments under Different Poses on the FERET Database}
In the FERET database, 600 samples of 200 subjects are chosen. Each subject provides three samples with a size of 20 $\times$ 20 pixels according to three different poses (front, left and right). Then, all 600 samples are utilized to construct the 2D data set $X$ and the wavelet transform [10] is performed twice on each sample in the data set $X$ to generate the corresponding 2D data set $Y$. Moreover, two experimental settings are adopted, front-left and front-right. In the first setting, the front samples are utilized for training while the left samples are for testing. In the second, the front images are still adopted as train samples but the right samples are utilized for testing. Again, since PCA, LDA, 2DPCA and 2DLDA are not able to explore the correlation between the two variable sets, they are applied to the data set $X$ only. On the other hand, 2DCCA, L2DCCA and CDTRL are performed on the 2D data sets $X$ and $Y$ while samples are reshaped into one dimensional vectors for CCA and DCCA. The experimental results are reported in TABLE II, demonstrating better performance of the proposed CDTRL.\\
\vspace*{-10pt}
\begin{table}[h]
\normalsize
\renewcommand{\arraystretch}{0.8}
\caption{\normalsize{The recognition accuracy with different methods on FERET dataset}}
\setlength{\abovecaptionskip}{0pt}
\setlength{\belowcaptionskip}{10pt}
\centering
\tabcolsep 0.1in
\begin{tabular}{ccc}
\hline
\hline
Method & Front-Left & Front-Right \\
\hline
PCA [13]  &77.50\% &75.50\% \\
2DPCA [14]  &78.50\% &76.50\% \\
LDA [15] &67.00\% &65.50\% \\
2DLDA [16] &71.50\% &70.50\% \\
CCA [5] &72.50\% &68.50\% \\
DCCA [24]  &78.50\% &60.50\% \\
2DCCA [9]  &80.50\% &74.50\%\\
L2DCCA [17]  &79.50\% &75.00\%\\
\textbf{The proposed CDTRL}  &\textbf{83.00\%} &\textbf{78.00\%}\\
\hline
\hline
\end{tabular}
\end{table}

\section{Conclusion}
This letter presents a CDTRL method for linear correlation analysis of 2D data. The main contribution of this letter is to generate the complete discriminative tensor representations across 2D data sets. It is demonstrated that CDTRL is more efficient than 2DCCA and L2DCCA at exploring the linear discriminant correlation between 2D data sets. For the proposed CDTRL method, since 2D data samples are utilized as inputs directly instead of reshaping them into one-dimensional vectors, lower computational complexity is expected. Experimental results show the superiority of the proposed CDTRL method.\\\indent Moreover, one of the worthwhile extensions is to conduct further investigation on the kernelized version of CDTRL based on $(2D)^2KCCA$ [10] to address nonlinear problems in the 2D data representation learning.





\ifCLASSOPTIONcaptionsoff
  \newpage
\fi


\begin{thebibliography}{1}

\bibitem{IEEEhowto:kopka}
M. Federici, A. Dutta, P. Forre, N. Kushman, and Z. Akata. ``Learning Robust Representations via Multi-View Information Bottleneck." \emph{2020 International Conference on Learning Representations (Accept)}.
\bibitem{IEEEhowto:kopka}
L. Gao, L. Qi, E. Chen and L. Guan, ``Discriminative multiple canonical correlation analysis for information fusion." \emph{IEEE Trans. on Image Processing}, vol. 27, no. 4, pp. 1951-1965, 2018.
\bibitem{IEEEhowto:kopka}
X. Xing, K. Wang, T. Yan, and Z. Lv. ``Complete canonical correlation analysis with application to multi-view gait recognition." \emph{Pattern Recognition}, vol. 50, pp. 107--117, 2016.
\bibitem{IEEEhowto:kopka}
A. de Cheveigne, G.M. Di Liberto, D. Arzounian, D.D. Wong, J. Hjortkjar, S. Fuglsang, and L.C. Parra. ``Multiway canonical correlation analysis of brain data." \emph{NeuroImage}, vol. 186, pp. 728--740, 2019.
\bibitem{IEEEhowto:kopka}
X. Jing, S. Li, C. Lan, D. Zhang, J. Yang, and Q. Liu. ``Color image canonical correlation analysis for face feature extraction and recognition." \emph{Signal Processing}, vol. 91, no. 8, pp. 2132--2140, 2011.
\bibitem{IEEEhowto:kopka}
L. Gao, L. Qi, and L. Guan. ``Online behavioral analysis with application to emotion state identification." \emph{IEEE Intelligent Systems}, vol. 31, no. 5, pp. 32--39, 2016.
\bibitem{IEEEhowto:kopka}
D. Lin, V.D. Calhoun, and Y. Wang. ``Correspondence between fMRI and SNP data by group sparse canonical correlation analysis." \emph{Medical image analysis}, vol. 18, no. 6, pp. 891--902, 2014.
\bibitem{IEEEhowto:kopka}
N. Sun, Z. Ji, C. Zou, and L. Zhao. ``Two-dimensional canonical correlation analysis and its application in small sample size face recognition." \emph{Neural Computing and Applications}, vol. 19, no. 3, pp. 377--382, 2010.
\bibitem{IEEEhowto:kopka}
S.H. Lee and S. Choi. ``Two-dimensional canonical correlation analysis." \emph{IEEE Signal Process. Lett.}, vol. 14, no. 10,  pp. 735-738, 2007.
\bibitem{IEEEhowto:kopka}
X. Gao, S. Niu, and Q. Sun. ``Two-Directional Two-Dimensional Kernel Canonical Correlation Analysis." \emph{IEEE Signal Processing Letters}, vol. 26, no. 11, pp. 1578--1582, 2019.
\bibitem{IEEEhowto:kopka}
http://www2.ece.ohio-state.edu/~aleix/ARdatabase.html.
\bibitem{IEEEhowto:kopka}
P.J. Phillips, H. Moon, S.A. Rizvi, and P.J. Rauss,``The FERET evaluation methodology for face-recognition algorithms." \emph{IEEE Trans. Pattern Anal. Mach. Intell.}, vol. 22, no. 10, pp. 1090--1104, 2000.
\bibitem{IEEEhowto:kopka}
F. Kherif, and A. Latypova. ``Principal component analysis." \emph{Machine Learning}, pp. 209--225, 2020.
\bibitem{IEEEhowto:kopka}
J. Yang, D. Zhang, A. F. Frangi and Jing-yu Yang. ``Two-dimensional PCA: a new approach to appearance-based face representation and recognition." \emph{IEEE transactions on pattern analysis and machine intelligence}, vol. 26, no. 1, pp. 131-137, 2004.
\bibitem{IEEEhowto:kopka}
P. Deng, H. Wang, T. Li, S. Horng, and X. Zhu. ``Linear discriminant analysis guided by unsupervised ensemble learning." \emph{Information Sciences}, vol. 480, pp. 211--221, 2019.
\bibitem{IEEEhowto:kopka}
 J. Ye, R. Janardan and Q. Li. ``Two-dimensional linear discriminant analysis." \emph{In Advances in neural information processing systems}, pp. 1569-1576, 2005.
\bibitem{IEEEhowto:kopka}
H.X. Wang. ``Local Two-dimensional canonical correlation analysis." \emph{IEEE Signal Process. Lett.}, vol. 17, no. 11, pp. 921--924, 2010.
\bibitem{IEEEhowto:kopka}
L. Gao, and L. Guan. ``A Discriminant Two-Dimensional Canonical Correlation Analysis." \emph{2019 IEEE Canadian Conference of Electrical and Computer Engineering (CCECE)}, pp. 1--4, 2019.
\bibitem{IEEEhowto:kopka}
L. Gao, L. Qi, E. Chen, and L. Guan. ``A fisher discriminant framework based on Kernel Entropy Component Analysis for feature extraction and emotion recognition." \emph{2014 IEEE International Conference on Multimedia and Expo (ICME)}, pp. 1--6, 2014.
\bibitem{IEEEhowto:kopka}
D. Tao, Y. Guo, Y. Li, and X. Gao. ``Tensor rank preserving discriminant analysis for facial recognition." \emph{IEEE transactions on image processing}, vol. 27, no. 1, pp. 325-334, 2018.
\bibitem{IEEEhowto:kopka}
X. Yang, W. Liu, and W. Liu. ``Tensor Canonical Correlation Analysis Networks for Multi-view Remote Sensing Scene Recognition." \emph{IEEE Transactions on Knowledge and Data Engineering (Early Access)}, 2020.
\bibitem{IEEEhowto:kopka}
S. Yang, M. Wang, Z. Feng, Z. Liu, and R. Li. ``Deep sparse tensor filtering network for synthetic aperture radar images classification." \emph{IEEE transactions on neural networks and learning systems}, vol. 29, no. 8, pp. 3919--3924, 2018.
\bibitem{IEEEhowto:kopka}
M. Wang, K. Zhang, X. Pan, and S. Yang. ``Sparse tensor neighbor embedding based pan-sharpening via N-way block pursuit." \emph{Knowledge-Based Systems}, vol. 149, pp. 18--33, 2018.
\bibitem{IEEEhowto:kopka}
L. Gao, L. Qi, E. Chen, and L. Guan. ``Discriminative multiple canonical correlation analysis for information fusion." \emph{IEEE Transactions on Image Processing}, vol. 27, no. 4, pp. 1951--1965, 2018.
\bibitem{IEEEhowto:kopka}
L. Gao, R. Zhang, L. Qi, E. Chen, and L. Guan. ``The labeled multiple canonical correlation analysis for information fusion." \emph{IEEE Transactions on Multimedia}, vol. 21, no. 2, pp. 375--387, 2019.
\bibitem{IEEEhowto:kopka}
X. Yang, W. Liu, W. Liu, and D. Tao. ``A survey on canonical correlation analysis." \emph{IEEE Transactions on Knowledge and Data Engineering(Early Access)}, 2019.

\end{thebibliography}
\end{document}